\def\year{2022}\relax
\documentclass[letterpaper]{article} 
\usepackage[submission]{aaai25}  
\usepackage{times}  
\usepackage{helvet}  
\usepackage{courier}  
\usepackage[hyphens]{url}  
\usepackage{graphicx} 
\usepackage{natbib}  
\usepackage{caption} 
\DeclareCaptionStyle{ruled}{labelfont=normalfont,labelsep=colon,strut=off} 
\usepackage{algorithm}
\usepackage{algorithmic}

\usepackage{newfloat}
\usepackage{listings}
\floatstyle{ruled}
\newfloat{listing}{tb}{lst}{}
\floatname{listing}{Listing}

\title{GGS: Generalizable Gaussian Splatting for Lane Switching in \\ Autonomous Driving}

\author{
}

\usepackage{bibentry}

\usepackage{amsmath,amsfonts}

\usepackage{array}
\usepackage{subfigure}
\usepackage{makecell}

\usepackage{multirow}

\usepackage{subcaption}

\usepackage{booktabs} 
\usepackage{utfsym} 

\begin{document}

\maketitle

\begin{abstract}

We propose GGS, a Generalizable Gaussian Splatting method for Autonomous Driving which can achieve realistic rendering under large viewpoint changes. Previous generalizable 3D gaussian splatting methods are limited to rendering novel views that are very close to the original pair of images, which cannot handle large differences in viewpoint. Especially in autonomous driving scenarios, images are typically collected from a single lane. The limited training perspective makes rendering images of a different lane very challenging. To further improve the rendering capability of GGS under large viewpoint changes, 
we introduces a novel virtual lane generation module into GSS method
to enables high-quality lane switching even without a multi-lane dataset. Besides, we design a diffusion loss to supervise the generation of virtual lane image to further address the problem of lack of data in the virtual lanes. 
Finally, we also propose a depth refinement module to optimize depth estimation in the GSS model. Extensive validation of our method, compared to existing approaches, demonstrates state-of-the-art performance.

\end{abstract}

\section{Introduction}







Novel view synthesis is an essential task in the field of computer vision, with significant potential applications in autonomous driving \cite{yang2020surfelgan, wu2023mars, liu2023real, yang2023unisim,  yang2024unipad, yu2024sgd}, object detection, and digital human representations. To enhance the robustness of autonomous driving systems, it is imperative to establish a simulation environment for testing these systems effectively. However, the majority of existing datasets are limited to single-lane scenarios. This limitation presents significant challenges in inferring adjacent lane scenarios from the current viewpoint. If lane switching is not supported, the test samples provided to the autonomous driving simulation system will be incomplete, making it impossible to conduct better simulation testing and requiring a significant amount of data collection costs.

The methods based on NeRF \cite{mildenhall2021nerf, yang2023emernerf} often rely on lidar to better generate novel views in autonomous driving scenarios. READ \cite{li2023read} introduces a new rendering method that adopts a neural rendering approach different from NeRF. It learns neural descriptors of the original point cloud with explicit geometry to render images, instead of learning implicit geometry in NeRF methods. However, the efficiency of training and rendering using these methods is very low.

\begin{figure}[t]

\centering
\includegraphics[width=0.47\textwidth]{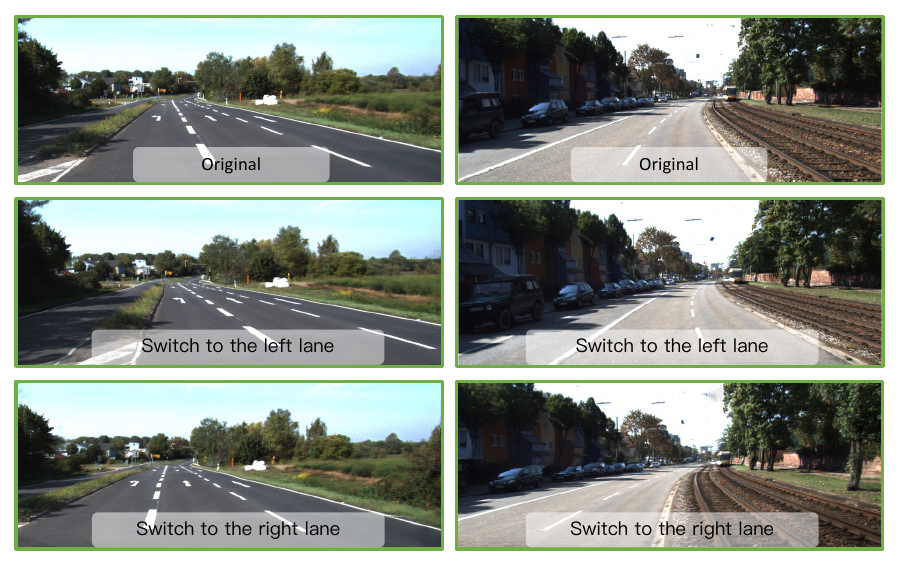} 
\caption{Our GGS method can achieve high-quality lane switching in autonomous driving scenarios.}
\label{fig:pic_first_page}

\end{figure}

The efficiency in training and rendering speed, coupled with the high reconstruction quality of 3D Gaussian Splatting \cite{kerbl20233d}, contributes to its widespread application of novel view synthesis in autonomous driving. GaussianPro \cite{cheng2024gaussianpro} introduces multi-view stereo to improve the geometry of generated gaussian splats. DC-Gaussian\cite{cheng2023uc} introduces an adaptive image decomposition module to address the impact of glass reflections on the quality of novel view synthesis. However, these methods still cannot have effective novel view synthesis in lane switching, as they do not address the main problem that only single lanes of data are collected.



To address the problem of the sparse view synthesis, many methods have sought to optimize this process using generative models \cite{yu2021pixelnerf, chen2024mvsplat, liu2024fast, wu2024reconfusion,tang2024mvdiffusion++}. generative models are trained across large amount of scenes to enhance performance in sparse view scenarios. However, the generative model still lacks of data from multi-lanes to learn how to synthesize novel views for other lanes from single lanes data. 

Therefore, we propose a virtual lane module into generative Gaussian splatting to address the synthesis of new views involving lane changes, despite the lack of multi-lane training datasets for supervision. In the module, we first use 3D gaussians generated from images in the single lanes using a generative model to predict images from virtual lanes, then use 3D gaussians generated from the virtual images to predict back the image collected in the single lanes. In this way, we can let generative model to learn how to generate the best images in the other lanes even with only single lanes of data. In addition, we propose a diffusion loss from a latent diffusion model \cite{sohl2015deep, song2020denoising, nichol2021improved, rombach2022high} to virtual generated images to further improve the lane switching of our GGS. Finally, as improving geometry of generated 3D gaussians also improves novel view synthesis in sparse view collections, we employs points from traditional multiview stereo reconstruction to refine the depth estimated in GGS.

The main contributions of this paper can be summarized as follows:

\begin{itemize}

\item We propose a novel virtual lanes module into the generative gaussian splatting to improve the quality of lane switching novel view with only single lanes of data.

\item We introduce a diffusion loss to directly supervise the image from virtual lanes predicted by GGS to further improve the novel view synthesis from limited collected views. 

\item We  propose to fusing MVS geometry into the generative 3D gaussian splating to improve geometry estimation.

\item We conduct extensive experiments on a wide range of scenarios to validate the effectiveness of our algorithm, and achieve state-of-the-art street novel view synthesis even without LiDAR.
\end{itemize}

\begin{figure*}[t]
    
\centering
\includegraphics[width=1.0\textwidth]{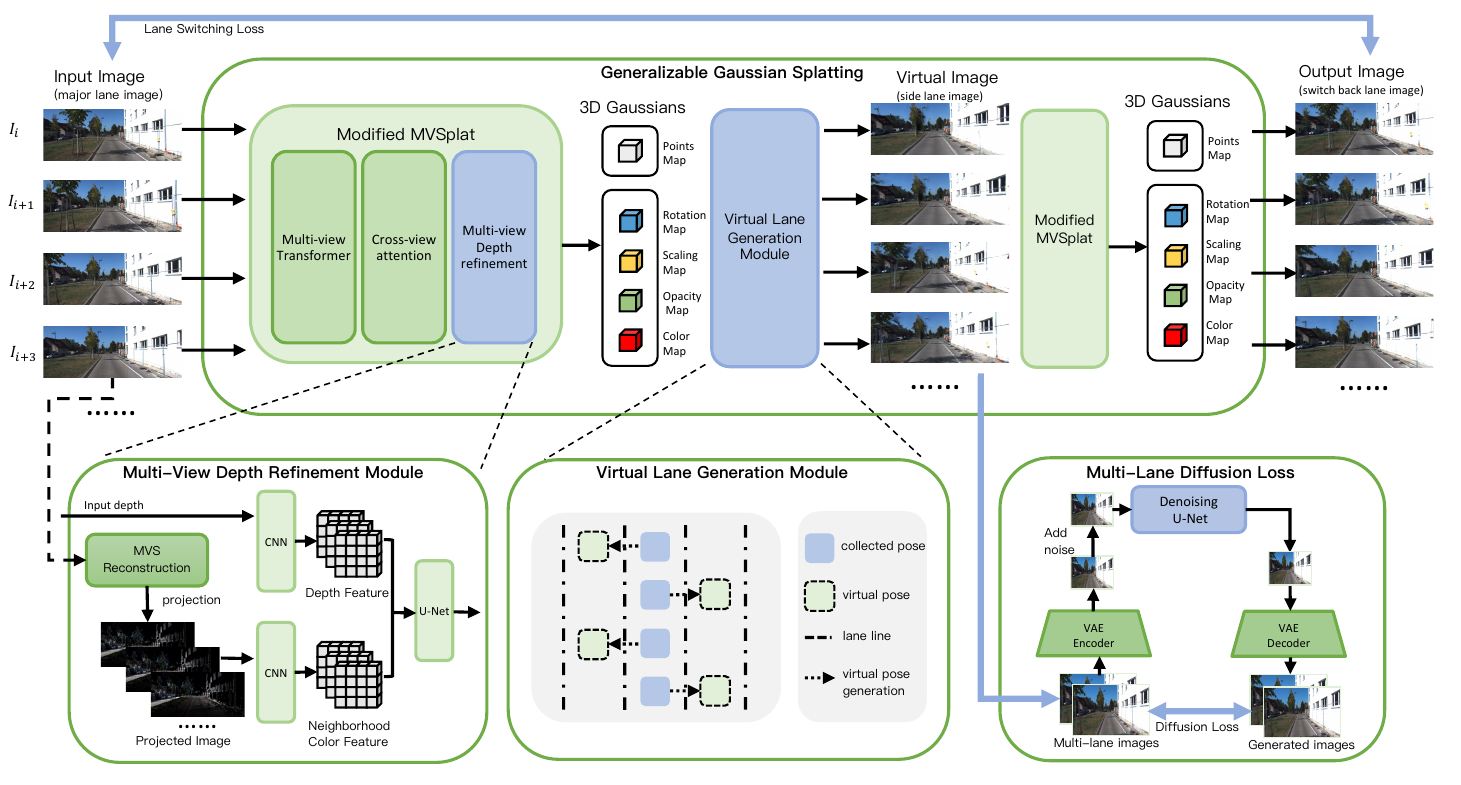} 


\caption{The overall framework of the GGS. Input multiple frames and estimate depth maps through MVS and multi view depth refinement modules, combined with 3DGS to synthesize novel views. And through the virtual lane generation module, switch lanes with high quality. In addition, multi-lane diffusion loss is introduced to supervise the novel view synthesis in the presence of obstacles.}

\label{fig:pic1}

\end{figure*}

\section{Related Work}

\subsection{3D Gaussian Splatting}
3D Gaussian Splatting \cite{kerbl20233d} employs a point-cloud-based 3D reconstruction method, which combines the position information of each point with Gaussian distribution to convert point cloud data into a 3D surface. However, the quality of street novel synthesis is still problematic due to limited view collections in the street.

GaussianPro \cite{cheng2024gaussianpro} has improved 3D Gaussian Splatting by introducing a novel progressive propagation strategy to guide Gaussian densification based on the scene's surface structure. Although improving geometries helps to mitigate the novel synthesis in sparse views, the quality of novel synthesis in another lanes is still low. Deformable 3D Gaussians \cite{yang2024deformable} employs a framework for extending 3D Gaussian Splatting in dynamic scenes using a deformation field, enabling the learning of 3D Gaussians in a normalized space. There are also other methods based on 3D Gaussian Splatting such as \cite{zhou2024drivinggaussian,paliwal2024coherentgs,niedermayr2024compressed}. Although street view synthesis has only been improved on the collected lanes, they have not solved the problem of sparse views, leaving lower levels of novel view synthesis when changing lanes.


\subsection{Generalized model}
To solve the problem of novel view synthesis in sparse views, some methods propose a generalized model-based approach. PixelNeRF \cite{yu2021pixelnerf} employs a generalized model for novel view synthesis based on volume rendering method, which can be trained directly from images without explicit 3D supervision. However, the generation quality is not high and the training efficiency is low. 
mvsplat \cite{chen2024mvsplat} introduces an efficient feedforward 3D Gaussian splash model learned from sparse multi view images, and constructs a cost volume to represent the cross view feature similarity of different candidate depths, providing valuable clues for depth estimation. MVSGaussian \cite{liu2024fast} employs a mixture Gaussian rendering method that integrates efficient volume rendering design for novel view synthesis. Compared with the original 3D Gaussian Splatting, MVSGaussian achieves better view synthesis results while reducing training computational costs. However, it cannot be well completed for scenes with obstacles.


\subsection{Diffusion Model}
For occluded scenes, using generalized models cannot generate better results, so some algorithms introduce diffusion model \cite{sohl2015deep,rombach2022high,nichol2021improved,song2020denoising} to imagine unknown regions. ReconFusion \cite{wu2024reconfusion} further utilizes the generative capacity of large models to infer unknown areas, and integrates diffusion prior into NeRF's 3D reconstruction process. DrivingDiffusion \cite{li2023drivingdiffusion} introduces a spatiotemporally consistent diffusion framework, incorporating multi-view attention to generate realistic multi-view videos controlled by 3D layouts. These diffusion model methods only consider a single lane and do not utilize multi-lane features for better completion.

\section{Methodology}




Although generalized models can assist in synthesizing novel views in sparse views, insufficient view information leads to inaccurate depth estimation. Our method further optimizes the generalization model. The overall framework diagram of our GGS method is as shown in Figure \ref{fig:pic1}. We input four different frame images and introduce neighborhood feature in the \textbf{Mutli-View Depth Refinement Module} to better address scenes with occlusions. And we introduce more global information to optimize the predicted depth map By using MVS. In the \textbf{Virtual Lane Generation Module}, we introduce the concept of virtual lanes and solve the problem of not having a multi-lane dataset by switching back after switching, allowing the model to flexibly switch lanes. In addition, we introduce the \textbf{Multi-Lane Diffusion Loss} to supervise the novel view synthesis.

\subsection{Background}

\textbf{MVSplat}~\cite{chen2024mvsplat} is a generalizable 3D Gaussian Splatting method, which can synthesize novel views from sparse inputs. MVSplat takes transformer based structure and adopts cross view attention strategy to build a cost volume for each input view, then following a U-Net to predict the depth and the parameters of Gaussian primitives for each pixel.
The 3D Gaussian parameters consist of the Gaussian center position $x$, scale $s$, rotation angle $q$, opacity $\alpha$, and color $c$. Given the predicted depth map $D$ and the projection matrix $P$ with camera parameters $K$, pixels located at $x$ are back-projected from the image plane to 3D space as follows:
\begin{equation}
x_{p_x} = \Pi_P^{-1}(p_x, D),
\end{equation}
where $\Pi$ represents the back-projection operation, and $p_x$ and $D$ represents pixel coordinates and estimated depth, respectively. The opacity $\alpha$ is represented by the matching confidence directly.
The remaining Gaussian parameters scale $s$, rotation angle $q$, and color $c$ are decoded from the encoded features as follows:
\begin{equation}    
s_{p_x} = Softplus(h_{s}(\Gamma(p_x))),
\end{equation}
\begin{equation}
q_{p_x} = Norm(h_{q}(\Gamma(p_x))),
\end{equation}
\begin{equation}
c_{p_x} = Sigmoid(h_c(\Gamma(p_x))),
\end{equation}
where $\Gamma$ represents the high-dimensional feature vector, $p_x$ represents pixel coordinates, and $h_s$, $h_q$, and $h_c$ represent the scaling head, rotation head, and color head, respectively.

\subsection{Mutli-View Depth Refinement Module}
We enhanced MVSplat by our Mutli-View Depth Refinement Module, i.e. Modified MVSplat. It can produce more accurate 3D gaussian primitives and improve the quality of novel view synthesis. To better infer unknown regions, we incorporated the color feature information of the neighborhood of this view. Our model takes this into consideration. We use the back-projected point cloud map reconstructed through Agisoft Metashape as an additional input color feature for U-Net. The feature representation of the neighborhood is:
\begin{equation}
F_{neighbor_{i}} = \{F_m | m \in [i-k, i+k]\},
\end{equation}
where $i$ represents the i-th frame in the video, and $F_i$ represents the color feature of the i-th frame. $k$ represents neighborhood distance.

Neighborhood color features are merged into depth features through concatenation, high-dimensional Gaussian parameter features are output through UNet, decoded using a Gaussian parameter decoder, and finally generate Gaussian parameter representations.


\begin{equation}
dep_{ref} = \mathcal U(F_{neighbor_{i}}, dep_{i}),
\end{equation}
where $\mathcal U$ represents the U-Net. By introducing color information from multiple neighborhood perspectives in this way, the synthesis ability of the generalized model under obstacle occlusion is enhanced.

In addition, to refine the depth, we introduce a confidence based method. The lower the transparency of the predicted 3D Gaussian, the lower the confidence level of the predicted depth. When the confidence level is high, the predicted depth remains unchanged. When the confidence level is low, we correct the predicted depth map by reconstructing the back-projected depth map through Agisoft Metashape \cite{metashape}. The optimized depth value is:
\begin{equation}
dep_{i} = \left\{\begin{matrix} 
\beta\hat{dep_{i}}+(1-\beta)D_{i},  if \space \alpha_{i}<\alpha \\  
\hat{dep_{i}},  if \space \alpha_{i}\ge \alpha
\end{matrix}\right.,
\end{equation}
where $D_{i}$ represents the depth of projected depth map. $\hat{dep_{i}}$ represents the predict depth. $\alpha$ and $\beta$ represent the transparency threshold and depth threshold, respectively.






\begin{figure}[t]

\centering
\includegraphics[width=0.4\textwidth,height=0.2\textwidth]
{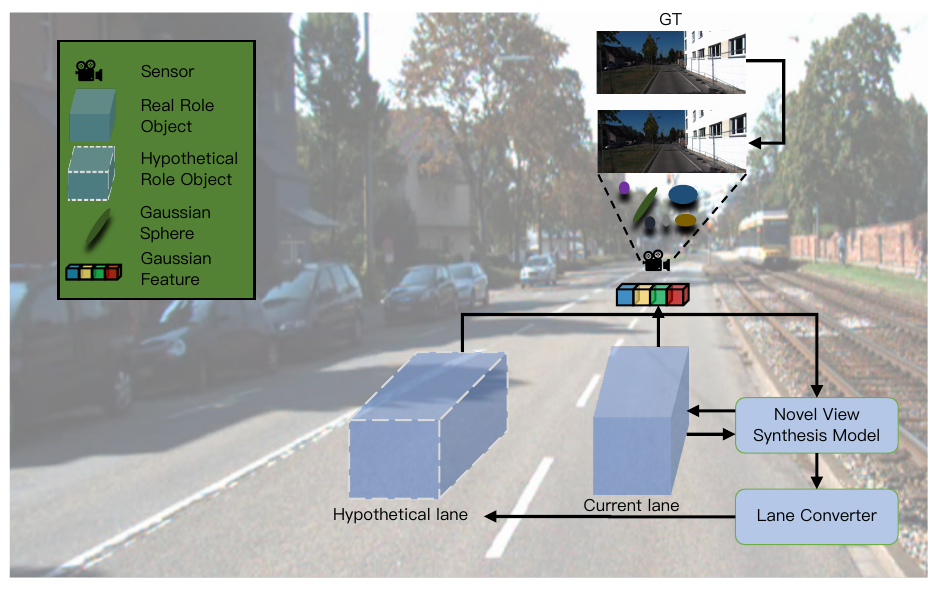} 
\caption{The method of using a lane converter to create a virtual lane and then switching back to the real lane enables the model to improve the quality of lane switching. }
\label{fig:pic2}

\end{figure}

\begin{figure}[t]


%
\centering
\includegraphics[width=0.45\textwidth]{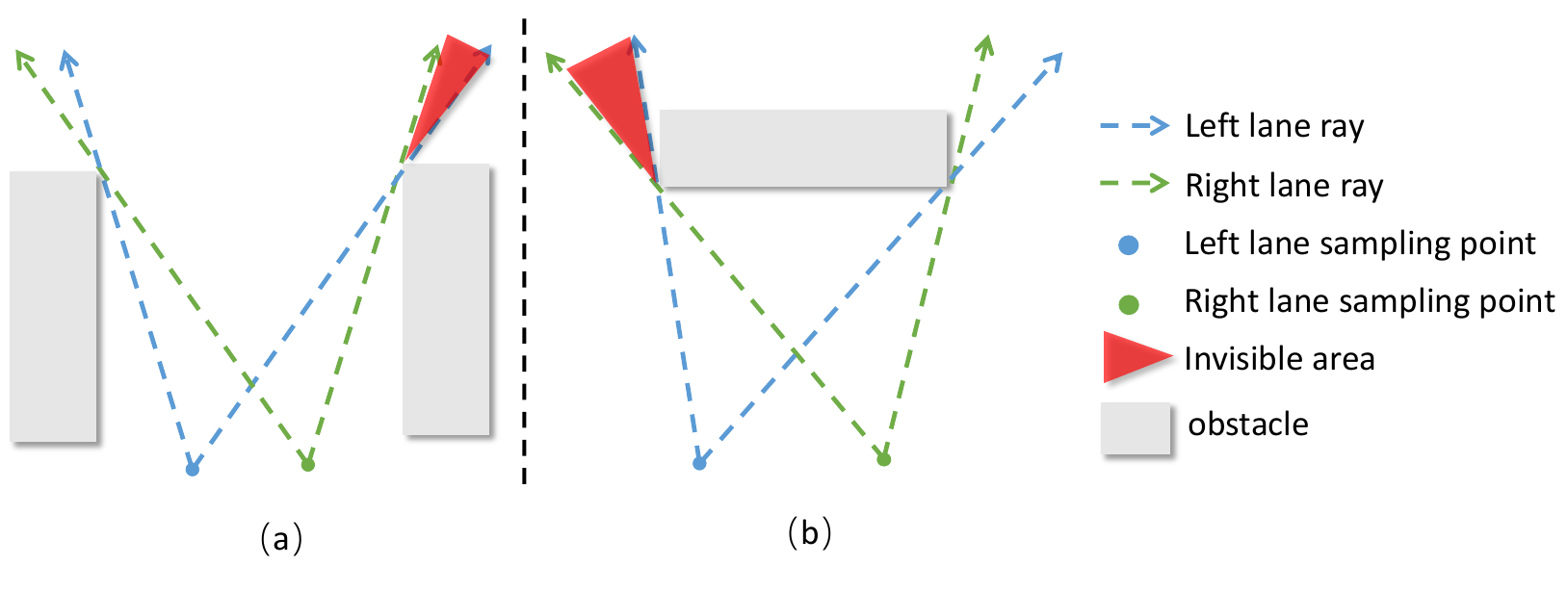} 
\caption{If we switch from the right lane to the left lane, the red area represents the blind spot of the right lane. When rendering the left lane, in order to avoid voids, the content of that area needs to be imagined in some way. }

\label{fig:pic_obstacle}

\end{figure}



\begin{figure*}[t]

\centering

\includegraphics[width=0.96\textwidth,height=0.65\textwidth]{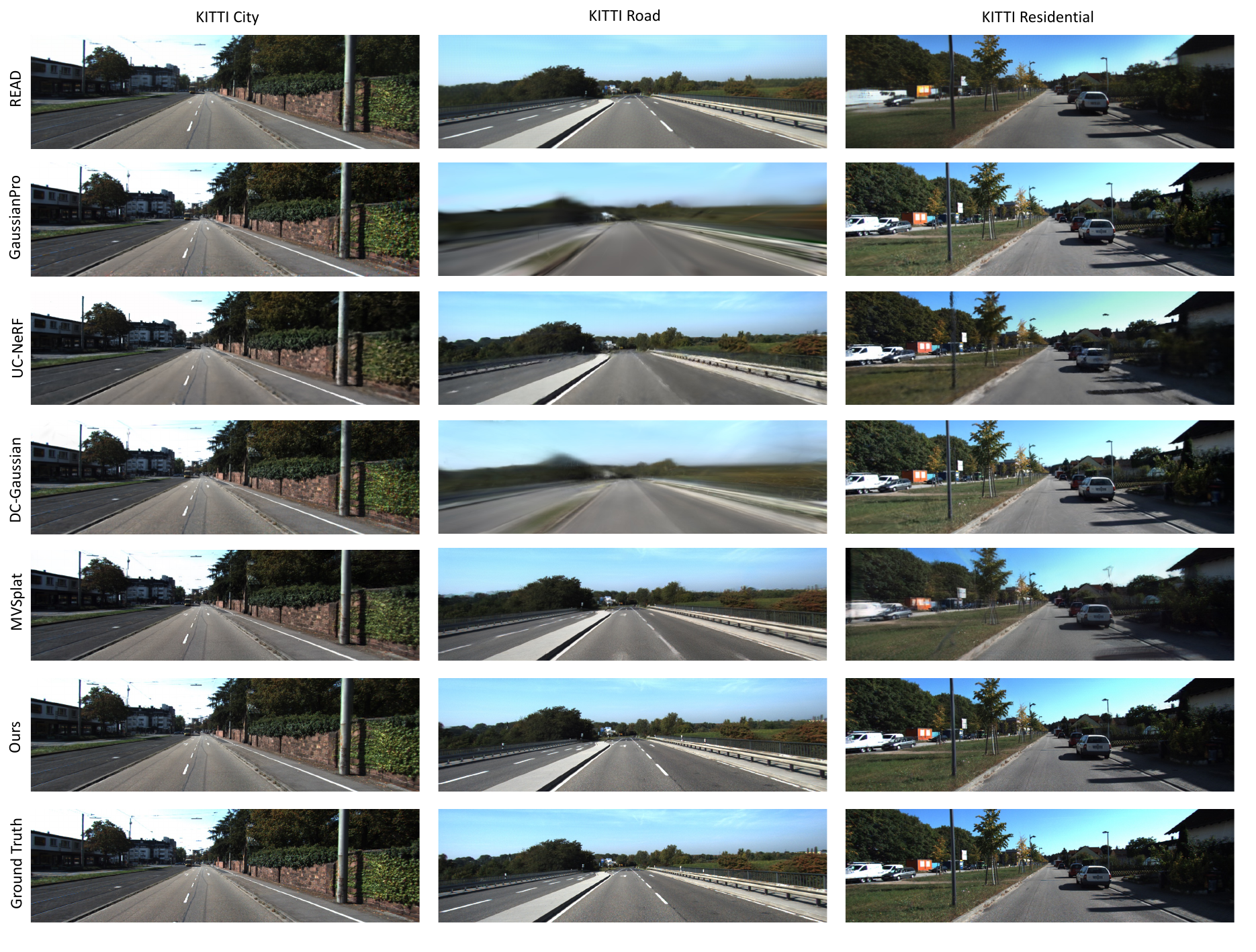} 
\caption{
Comparison results of novel view synthesis based on KITTI for residential, road, and urban scenes.
}
\label{fig:test_pic1}

\end{figure*}


\begin{figure*}[t]
\centering
\includegraphics[width=0.96\textwidth,height=0.55\textwidth]{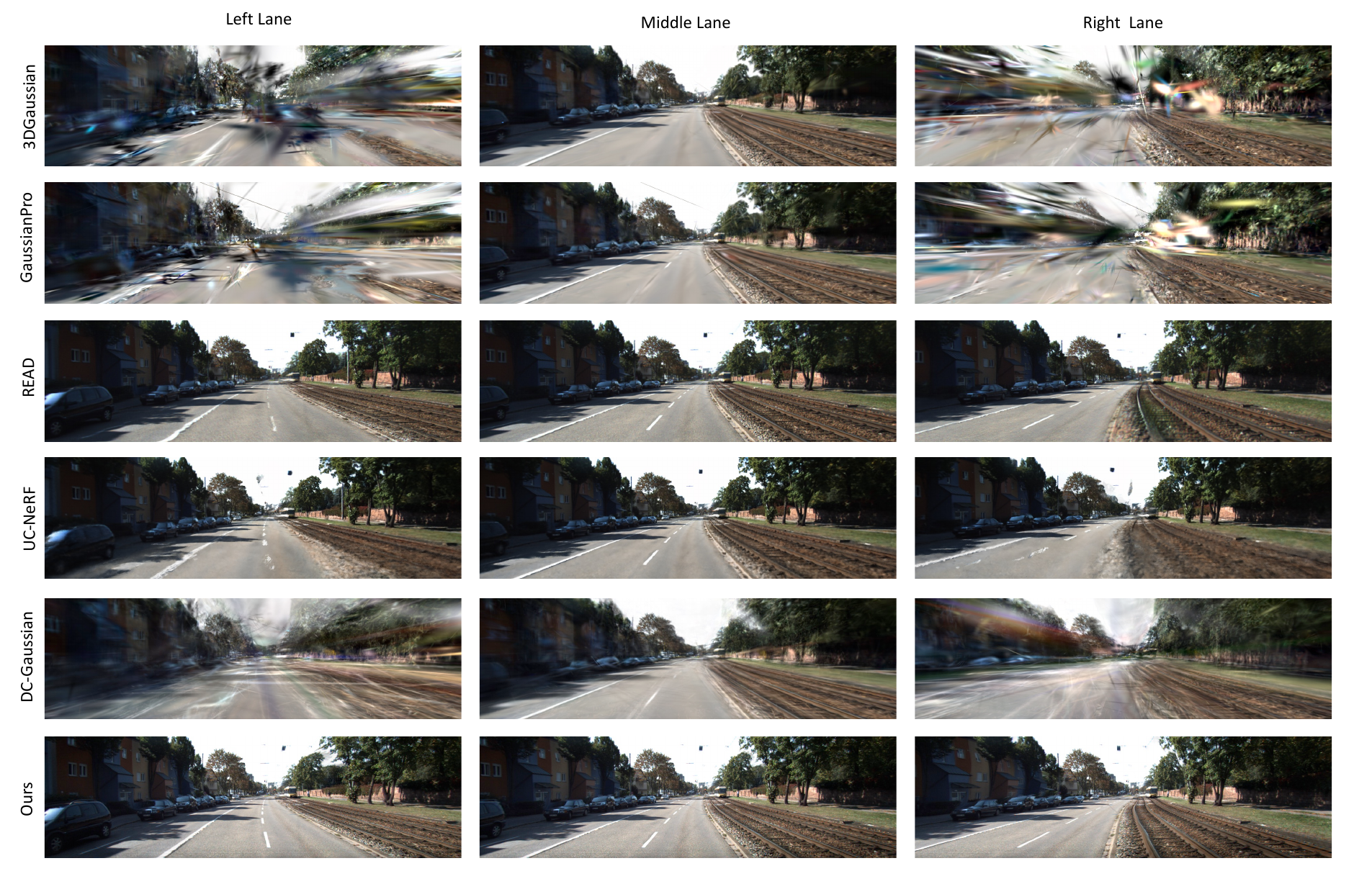} 
\caption{
Comparison of lane switching between different
models on KITTI dataset.
}
\label{fig:test_pic2}

\end{figure*}

\subsection{Virtual Lane Generation Module}


Previous generalizable 3D gaussian splatting methods are limited to rendering novel views that are very close to the original pair of images, which cannot handle large differences in viewpoint. Especially in autonomous driving scenarios, images are typically collected from a single lane. The limited training perspective makes rendering images of a different lane very challenging. We have obtained a 3D Gaussian using our modified depth refinement module. To further improve the rendering capability of GGS under large viewpoint changes, We introduce the virtual lane approach that enables high-quality lane switching even without a multi-lane dataset,  inspired by \cite{huang2023neural}. 


The virtual lane converter is used to select the appropriate virtual lane, so that after lane switching, no information can be seen from the virtual perspective due to excessive switching amplitude. Then generate a pose for the virtual lane by performing a vertical translation along the lane. Finally, generate a virtual perspective based on the pose of the virtual lane. After introducing the virtual lane module, our GGS module process mainly includes two stages.

In the first stage, we input a set of N images:
\begin{equation}
ISet_1 = \{I_1, I_2, ...I_N\},
\end{equation}
then we output the target image through the model:
\begin{equation}
\hat{I^1}= \mathcal G(ISet_1),
\end{equation}
where $\mathcal G$ represents GGS module, and $ISet_1$ is a rendered image without shifting the view, and the rendered view is consistent with the ground truth. The current lane generates a collection of virtual lane rendering images through lane converters. The rendered image of the virtual lane is represented as:
\begin{equation}
{ISet_{2} = \{\mathcal V(\hat{I^1_k},\gamma sin\theta)|k_f\le k\le k_l, \theta=\omega k\}},
\end{equation}
where $\mathcal V$ represents the virtual lane converter, $\gamma$ represents the translation coefficient, $k_f$ and $k_l$ represent the index of the first and last frames of the input, respectively. $\omega$ represents the switching period angle, and the switching angle of each frame changes periodically in order.




In the second stage, use the virtual lane generated in the first stage as input. Using our model, switch back from the virtual lane to the real lane and output a rendered image of the real lane:
\begin{equation}
\hat{I^2}= \mathcal G(ISet_2),
\end{equation}
where $\mathcal G$ represents GGS module. This forms a closed-loop process of switching to a new lane and then switching back. The advantage of doing so is that even without the ground truth of the left and right lanes, we can still enhance the quality of the model's rendering of the left and right lanes by establishing virtual lanes, allowing the model to improve the quality of lane switching, as shown in Figure \ref{fig:pic2}.

\subsection{Multi-Lane Diffusion Loss}

There is no ground truth available for training when switching lanes. When the lane switching amplitude is large, obstacles can obstruct the view during lane changes, making it impossible to collect information about the new lane from the current lane, as shown in Figure \ref{fig:pic_obstacle}. Therefore, in order to better address this issue, we use diffusion prior knowledge to imagine color information from a novel lane view. 

The traditional diffusion model denoising method directly completes the generated image, but due to the diversity of the generated models, it can lead to inconsistent results between frames. Our method calculates the loss of the denoised image and the image before denoising, and generates a new perspective supervised by diffusion. Additionally we construct multi-lane novel view images, instead of utilizing image of the current lane as input for U-Net denoising. This approach helps ensure that the autonomous driving lane remains visible in the image following a change in viewpoint.

Specifically, we adapt the Stable Diffusion framework \cite{rombach2022high}, and use the Variational AutoEncoder \cite{kingma2013auto} to encode the multi-lane images into latent code, including the left lane, middle lane and right lane. Then, we perform several denoising steps on the latent code as an initialization parameter for Denoising U-Net, fixing the input text as the autonomous driving label. Generated through the CLIP \cite{radford2021learning}, denoised through several steps, and then decoded into images using the Variational AutoEncoder. These images serve as supervision to guide the synthesis of novel views.

\subsection{Loss Function}

Our model is trained on a single lane dataset and introduces a method of constructing virtual lanes to generate unknown domains through diffusion models. Therefore, our method mainly includes reconstruction loss, depth loss, virtual lane switching loss, and diffusion loss. The overall loss function is represented as follows:
\begin{equation}
\mathcal{L}=\mathcal{L}_{\text {recon }}+\mathcal{L}_{\text {depth }}+\mathcal{L}_{\text {switch }}+\mathcal{L}_{\text {diffusion }}.
\end{equation}

\begin{table*}[htbp]
\renewcommand\arraystretch{0.8} 
\setlength{\tabcolsep}{5.5pt} 
\small
\begin{center}
\caption{Quantitative evaluation of novel view synthesis on KITTI dataset and Brno dataset.}

\label{table:table1}

\begin{tabular}{ c | c  c  c  c | c  c  c  c | c  c  c  c }
\toprule[1.2pt]
 & \multicolumn{4}{c|}{KITTI Residential} & \multicolumn{4}{c|}{KITTI Road} & \multicolumn{4}{c}{KITTI City}\\
 
& VGG$\downarrow$ &	PSNR$\uparrow$ & LPIPS$\downarrow$ &  	 SSIM$\uparrow$ & VGG$\downarrow$ &	PSNR$\uparrow$ & LPIPS$\downarrow$ &  	 SSIM$\uparrow$ & VGG$\downarrow$ &	PSNR$\uparrow$ & LPIPS$\downarrow$ &  	 SSIM$\uparrow$ \\ 
 
\hline
\multicolumn{13}{c}{Test on KITTI dataset} \\

\hline 
ADOP &	610.8 & 19.07 & 0.2116 & 0.5659 & 577.7 & 19.67 & 0.2150 & 0.5554 & 560.9 & 20.08 & 0.1825 & 0.6234 \\
READ &	454.9 & 22.09  &0.1755  &0.7242  &\textbf{368.2}  &24.29  &\textbf{0.1465}  &0.7402 & 391.1  &23.48  &0.1321 & 0.7871\\
UC-NeRF &555.1	&23.7	&0.4229	&0.7564		
&772.9	&20.62	&0.4998	&0.6502		
&469.2	&24.7	&0.3453	&0.7555 \\
3DGaussian &585.8	&22.66	&0.3683	&0.7859		
&760	&20.92	&0.4544	&0.7331		
&372.4	&24.92	&0.2258	&0.8566    \\
GaussianPro & 532.5	&23.74	&0.337	&0.8006		
&602.5	&23.46	&0.3803	&0.78		
&327.9	&24.84	&0.1999	&0.8763   \\
DC-Gaussian &416.6	&25.63	&0.2739	&0.8406		
&707.4	&21.28	&0.417	&0.7422		
&343.4	&25.04	&0.2115	&0.8713  \\
MVSplat &549.6	&22.45	&0.3008	&0.6562		&515.6	&21.08	&0.2749	&0.7457		&369.9	&24.11	&0.1667	&0.7755 \\
Ours & \textbf{259.1}	&\textbf{26.26}	&\textbf{0.0948}	&\textbf{0.8840}		
&372.6	&\textbf{25.01}	&0.1542	&\textbf{0.827}		
&\textbf{271.6}	&\textbf{26.79}	&\textbf{0.0933}	&\textbf{0.8781} \\

\hline
\multicolumn{13}{c}{Test on Brno Urban dataset} \\

\hline 
 & \multicolumn{4}{c|}{Left side view} & \multicolumn{4}{c|}{Left front side view} & \multicolumn{4}{c}{Right side view}\\
 \hline 
ADOP &	634.0 &19.19 &0.2414 &0.5927 &520.6 &20.83 &0.2189 &0.6633 &807.1 &16.51 &0.3636 &0.5009 \\
READ &	459.8 &21.79 &0.1905 &0.7067 &\textbf{341.1} &24.85 &\textbf{0.1513} &0.7836 &663.6 &18.44 &0.3065 &0.5771 \\
UC-NeRF &640.9 & 23.47 & 0.5201 & 0.8318
 & 900.7 & 20.28 & 0.6315 & 0.7251
 & 431.8 & 27.27 & 0.4212 & 0.7977 \\
3DGaussian &530.6 & 25.63 & 0.3583 & 0.828
 & 753.5 & 19.05 & 0.5634 & 0.7675
 & 400.8 & 28.02 & 0.3229 & 0.8629 \\
GaussianPro&520.5 & 25.75 & 0.3501 & 0.8307
 & 738.6 & 19.43& 0.5528 & 0.7731
 & 394.0 & \textbf{28.27}& 0.3151 & 0.8623 \\
DC-Gaussian & 699.1 & 20.86 & 0.4772 & 0.8102
 & 493.5 & 25.82 & 0.3373 & 0.8404
 & 303.4 & 26.95& 0.2692 & \textbf{0.898} \\
MVSplat &383.3	&24.97	&0.1652	&0.8013		&513.7	&22.49	&0.2712	&0.7693		&511.4	&22.21	&0.3806	&0.6575 \\
Ours &\textbf{275.3}   & \textbf{27.8}     & \textbf{0.0861}  & \textbf{0.8829}  
 & 354.7 & \textbf{25.85} & 0.1679  & \textbf{0.8415}
 & \textbf{288.9} & 26.5 & \textbf{0.1833} & 0.8936 \\

\bottomrule[1.2pt]
\end{tabular}
\end{center}
\end{table*}

\begin{table}[htbp]
\renewcommand\arraystretch{0.8}
 \setlength{\tabcolsep}{1.06pt} 
\small
\begin{center}
\caption{Ablation study on KITTI dataset.}

\label{table:table3}

\begin{tabular}{ c | c  c  c  c }
\toprule[1.2pt]

 & VGG$\downarrow$ & PSNR$\uparrow$ & LPIPS$\downarrow$ & SSIM$\uparrow$  \\ 
\hline

baseline & 	398.1 & 25.14 & 0.1935 & 0.7665 \\
w/o virtual lane generation module &257.7 & 26.76 & 0.0783 & 0.8808 \\
w/o multi-lane diffusion loss  & 215.4 & 28.97 & 0.0681 & 0.9056 \\
w/o depth refinement module & 213.5 & 29.05 & 0.0670 & 0.9074 \\
Ours(full model)& 	\textbf{210.8} & \textbf{29.12} & \textbf{0.0657} & \textbf{0.9087} \\

\bottomrule[1.2pt]
\end{tabular}
\end{center}
\end{table}


\begin{table}[t]
\renewcommand\arraystretch{0.40}
\setlength{\tabcolsep}{14.6pt} 

\footnotesize
\begin{center}
\caption{Lane switching experiment on KITTI dataset.}

\label{table:table4}

\begin{tabular}{ c | c  c  c }
\toprule[1.2pt]
  & FID@LEFT$\downarrow$  & FID@RIGHT$\downarrow$ \\ 
\hline


3DGaussian  &  201.13  & 164.01   \\
GaussianPro & 	198.35  & 154.94  \\
DC-Gaussian & 165.13  & 209.39  \\
UC-NeRF & 	106.89  & 82.98 \\
READ & 	79.79 & 76.87 \\
Ours& \textbf{60.34}  & \textbf{55.17}  \\
\bottomrule[1.2pt]
\end{tabular}
\end{center}
\end{table}

\textbf{Reconstruction loss}. Our GGS model is a generative model for novel views on autonomous driving. During the training process, we construct a reconstruction loss function by comparing the rendered image with the ground truth using mean square error loss.
\begin{equation}
\mathcal{L}_{\text {recon }}=\frac{1}{n}\sum_{i=1}^{n}\left(y_{i}-\hat{y}_{i}\right)^{2},
\end{equation}
where $y_{i}$ represents the color value in Ground Truth corresponding to a certain pixel, and $\hat{y}_{i})$ represents the color value in the rendered image corresponding to the same pixel.



\begin{figure}[ht]

\centering
\includegraphics[width=0.48\textwidth,height=0.3\textwidth]{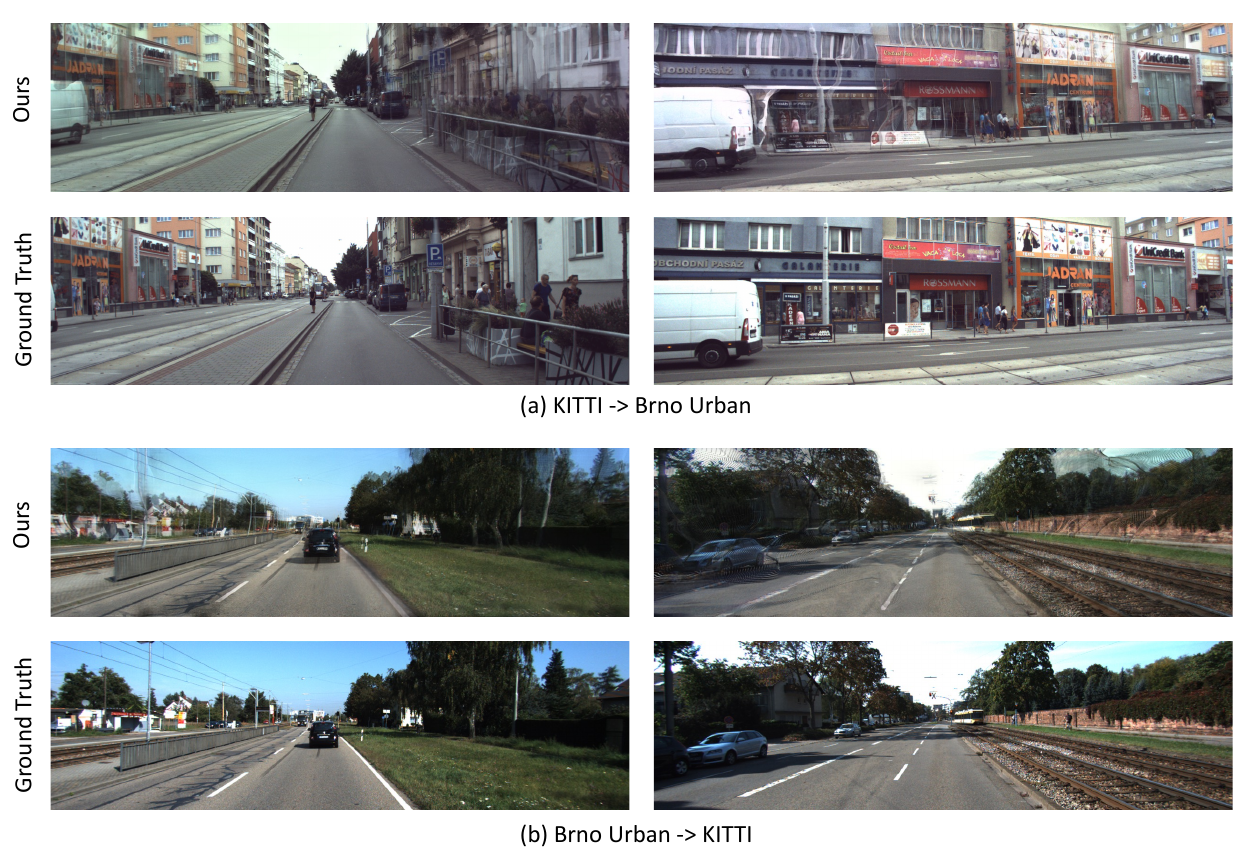} 
\caption{
Cross-dataset generalization. (a) train the model on the KITTI dataset and test it on the Brno Urban dataset. (b) train the model on the Brno Urban dataset and test it on the KITTI dataset.
}
\label{fig:test_generalization}

\end{figure}

\textbf{Depth loss}. In most autonomous driving scenarios, lanes are regular and even, so the depth of adjacent pixels should be smooth to avoid abrupt changes. Therefore, we construct the depth loss function as follows:
\begin{equation}
\mathcal{L}_{\text {depth }}=\frac{1}{n}\sum_{i=1}^{n} (\frac{dD_{i}}{dx} +\frac{dD_{i}}{dy} + \lambda(\frac{d^{2}D_{i}}{dx^{2}} +\frac{d^{2}D_{i}}{dy^{2}})), 
\end{equation}
where $\frac{dD_{i}}{dx}$, $\frac{dD_{i}}{dy}$, $\frac{d^{2}D_{i}}{dx^{2}}$ and $\frac{d^{2}D_{i}}{dy^{2}}$ represents the first and second derivatives of the depth in the x and y-axis directions of the image, respectively. And $\lambda$ is the depth smoothing adjustment factor.

\textbf{Lane switching loss}. Due to the lack of lane switching data, we train the model by constructing virtual lanes and switching back, and construct a lane switching loss.
\begin{equation}
\mathcal{L}_{\text {switch }}=\frac{1}{n}\sum_{i=1}^{n}\left(y_{i}-\Psi(\Phi (\hat{y}_{i}))\right)^{2},
\end{equation}
where $\Phi$ represents constructing virtual lanes and $\Psi$ represents switching from the virtual lane to the current lane.

\textbf{Multi-lane diffusion loss}. When we switch lanes in autonomous driving, changes in view can cause artifacts, so we use denoising methods to eliminate noise. 

\begin{equation}
\mathcal{L}_{\text {diffusion }}=\mathbb{E}_{\pi, t}\left[\beta(t) \left(\left\|y-\hat{y}_{\pi}\right\|_{1}+\mathcal{L}_{\mathrm{lpips}}\left(y, \hat{y}_{\pi}\right)\right)\right],
\end{equation}
where $\pi$ represents the camera pose of the selected views, $y$ represents the multi-lane images, $\hat{y}_{\pi}$ represents the output images from the denoising model, $\beta(t)$ is a weight function related to the noise level, and $\mathcal{L}_{\mathrm{lpips}}$ represents perceptual loss, which aims to emulate human perception of image similarity to better capture visual differences between images.

\section{Experiments}



We compare GGS with ADOP \cite{ruckert2022adop}, READ \cite{li2023read}, 3DGaussian \cite{kerbl20233d}, GaussianPro \cite{cheng2024gaussianpro}, UC-NeRF \cite{cheng2023uc} and DC-Gaussian \cite{wang2024dc}. We use Peak Signal-to-Noise Ratio (PSNR), Structural Similarity Index (SSIM), perceptual loss (VGG loss), perceptual metrics, and Learned Perceptual Image Patch Similarity (LPIPS) as evaluation metrics. 

\subsection{Evaluation on KITTI and BrnoUrban}

From Table \ref{table:table1}, methods based on 3D Gaussian Splatting, such as GaussianPro, and DC-Gaussian, generate slightly better quality than other methods based on neural radiation fields. However, in some scenes, the rendering quality is inferior, and our model performs better.


As illustrated in Figure \ref{fig:test_pic1}, GaussainPro and DC-Gaussian fail to capture details such as tree leaves and utility poles. The rendering quality of the READ is inadequate, and UC-NeRF does not render the white lines in the middle of the road. The comparison methods of different models for lane switching are shown in Figure \ref{fig:test_pic2}. Compared to other models, our method demonstrates excellent overall rendering quality and lane switching quality. 

\subsection{Assessing Cross-dataset Generalization}
Our method GGS has the advantage of generalization in extending to new scenarios outside the distribution. To evaluate the generalization of our model, we conduct two cross-dataset evaluations. Specifically, we train the model on KITTI dataset and test it on Brno Urban dataset \cite{ligocki2020brno}. Conversely, we train the model on Brno Urban and test it on KITTI, as shown in Figure \ref{fig:test_generalization}.

\begin{figure}[t]
\centering
\includegraphics[width=0.48\textwidth]{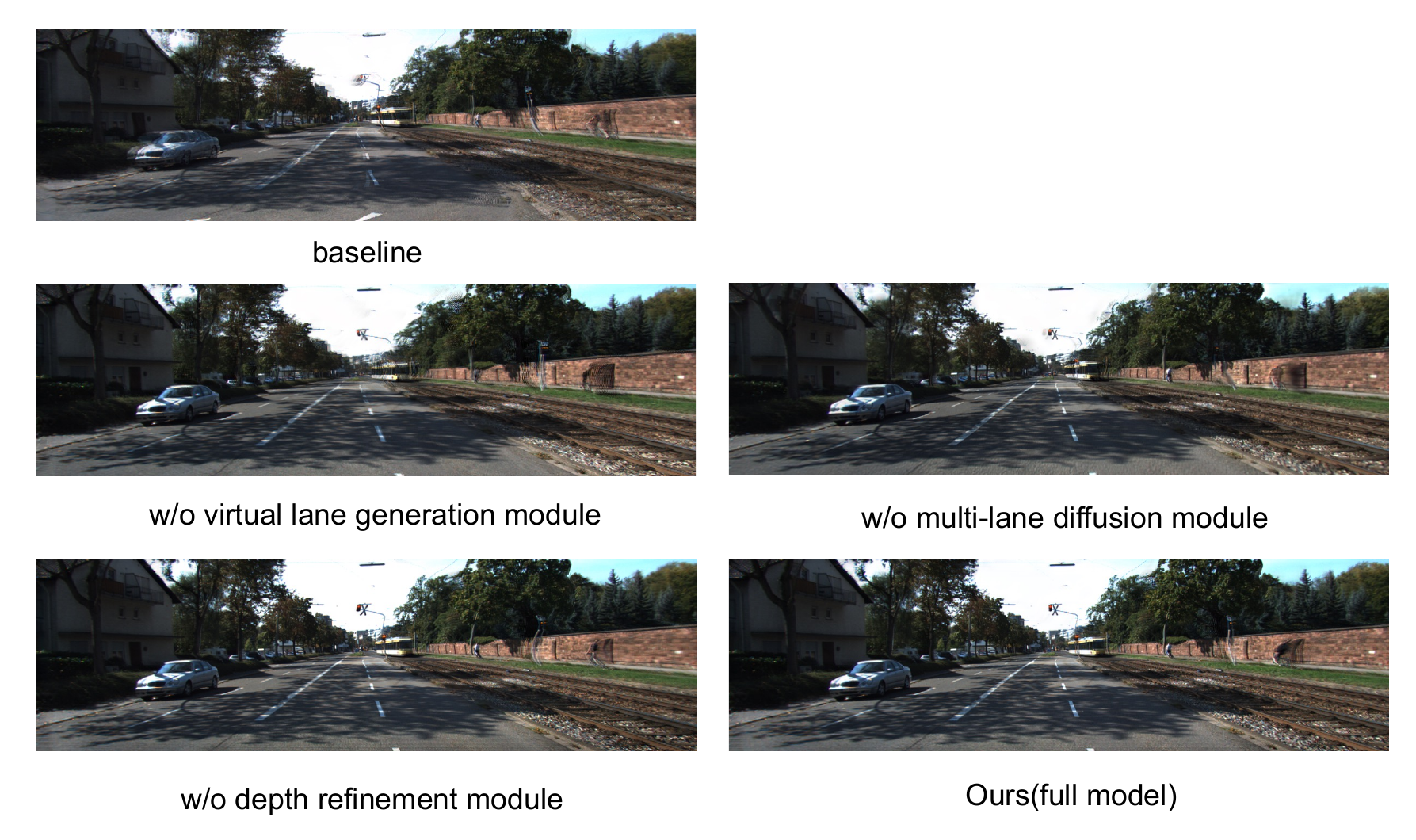} 
\caption{
Qualitative ablation study on KITTI dataset.
}
\label{fig:ablation}

\end{figure}

\subsection{Ablation Study}





\textbf{Effect of Virtual Lane Generation module}. To demonstrate the effectiveness of the virtual lane generation module, we use FID \cite{heusel2017gans} to conduct lane-switching experiments on different models, as shown in Table \ref{table:table4}. FID@LEFT and FID@RIGHT represents the distance between the rendered images of the left and right lanes and the GT. The qualitative experimental results are illustrated in Figure \ref{fig:test_pic2}. Our model achieves high rendering quality while ensuring that quality remains unaffected during lane switching, with quantitative results shown in Table \ref{table:table3} and qualitative results shown in Figure \ref{fig:ablation}.

\textbf{Effect of Multi-Lane Diffusion Loss}. Due to limited input view information, some unknown areas cannot be synthesized after lane switching. Therefore, a diffusion model is used to imagine the unknown areas and optimize the generation quality, as shown in Table \ref{table:table3}.


\textbf{Effect of Depth Refinement Module}. The depth refinement module introduces neighborhood feature information to optimize depth estimation in the presence of occluded objects, as shown in Table \ref{table:table3}. After removing the deep refinement module, each metrics affected slightly.

\section{Conclusions}


In this paper, we have proposed a generative framework based on MVS and 3D Gaussian Splatting fusion, which can repair unknown regions to optimize generation quality. By simulating the virtual lanes, our method effectively switches driving lanes in autonomous driving scenarios, suitable for simulation testing of autonomous driving systems. Our method has some limitations, the quality of lane switching generation needs to be improved when encountering dynamic scenes with complex road conditions, multiple people, and mixed vehicles.

\bibliography{main}


\end{document}